%% file: main.tex
\documentclass[11pt]{article}

\usepackage[T1]{fontenc}
\usepackage[utf8]{inputenc}
\usepackage{arabtex}
\usepackage{utf8}  %
\setcode{utf8}

\usepackage[preprint]{acl}

\usepackage{times}
\usepackage{latexsym}

\usepackage{microtype}

\usepackage[noupquote]{inconsolata}

\usepackage{graphicx}

\usepackage{booktabs}
\usepackage{amsmath}
\usepackage{amssymb}
\usepackage{dsfont}  %
\usepackage{siunitx}
\usepackage{array}
\usepackage{multirow}
\usepackage{CJKutf8}

\title{LatentMT: Machine Translation with Latent Reasoning}

\author{
 \textbf{Wei-Rui Chen\textsuperscript{1,*}},
 \textbf{Samar M. Magdy\textsuperscript{1}},
 \textbf{Chiyu Zhang\textsuperscript{1}},
 \textbf{Wenhui Zhu\textsuperscript{2}},
\\
 \textbf{Zhipeng Wang\textsuperscript{3}},
 \textbf{Muhammad Abdul-Mageed\textsuperscript{1,4,*}},
\\
\\
 \textsuperscript{1}The University of British Columbia,
 \textsuperscript{2}Arizona State University, \\
 \textsuperscript{3}Rice University,
 \textsuperscript{4}Canada Research Chair in NLP and ML
\\
 \small{
   \texttt{wrchen012@gmail.com, muhammad.mageed@ubc.ca}
 }
}

\begin{document}
\maketitle

\begingroup\def\thefootnote{*}\footnotetext{Corresponding authors}\endgroup

\begin{abstract}
Latent-reasoning looped language models (LoopLMs) offer a different scaling path for machine translation (MT): instead of increasing parameter count or emitting explicit chain-of-thought tokens, they spend additional recurrent computation inside hidden states.
We introduce \textbf{LatentMT}, the first systematic study of latent-reasoning LoopLMs for machine translation.
LatentMT adapts a small 2.6B-parameter backbone model with lightweight training.
Across 32 translation directions spanning high-, mid-, and low-resource languages, LatentMT achieves performance comparable to models three to five times larger.
It is competitive in a high-resource language and achieves state-of-the-art performance on both mid-resource and low-resource languages.
Studying the behavior of scaling the number of recurrent reasoning steps, we find that recurrent computation consistently improves translation quality in early steps, then saturates quickly afterwards.
Our mechanistic analysis shows that hidden-representation differences shrink along the recurrent reasoning-step axis, supporting the observed saturation in performance.
Finally, our efficiency analysis shows that LatentMT requires lower training and inference compute than much larger non-latent-reasoning models with similar performance, making latent recurrent computation a promising path toward compact, efficient, and strong machine translation.
\end{abstract}

\input{section/intro}
\input{section/related}
\input{section/experiments}
\input{section/results}
\input{section/analysis2}
\input{section/conclusion}

\section*{Acknowledgments}
Muhammad Abdul-Mageed acknowledges support from Canada Research Chairs (CRC), the Natural Sciences and Engineering Research Council of Canada (NSERC; RGPIN-2026-07098), the Social Sciences and Humanities Research Council of Canada (SSHRC; 895-2020-1004), Canadian Foundation for Innovation (CFI; 37771), Digital Research Alliance of Canada\footnote{\href{https://alliancecan.ca}{https://alliancecan.ca}}, UBC Advanced Research Computing-Sockeye\footnote{\href{https://arc.ubc.ca/ubc-arc-sockeye}{https://arc.ubc.ca/ubc-arc-sockeye}}.

\bibliography{custom}

\appendix

\input{section/appendix}

\end{document}

%% file: section/intro.tex
\section{Introduction}

\begin{figure}[t]
	\centering
	\includegraphics[width=\columnwidth]{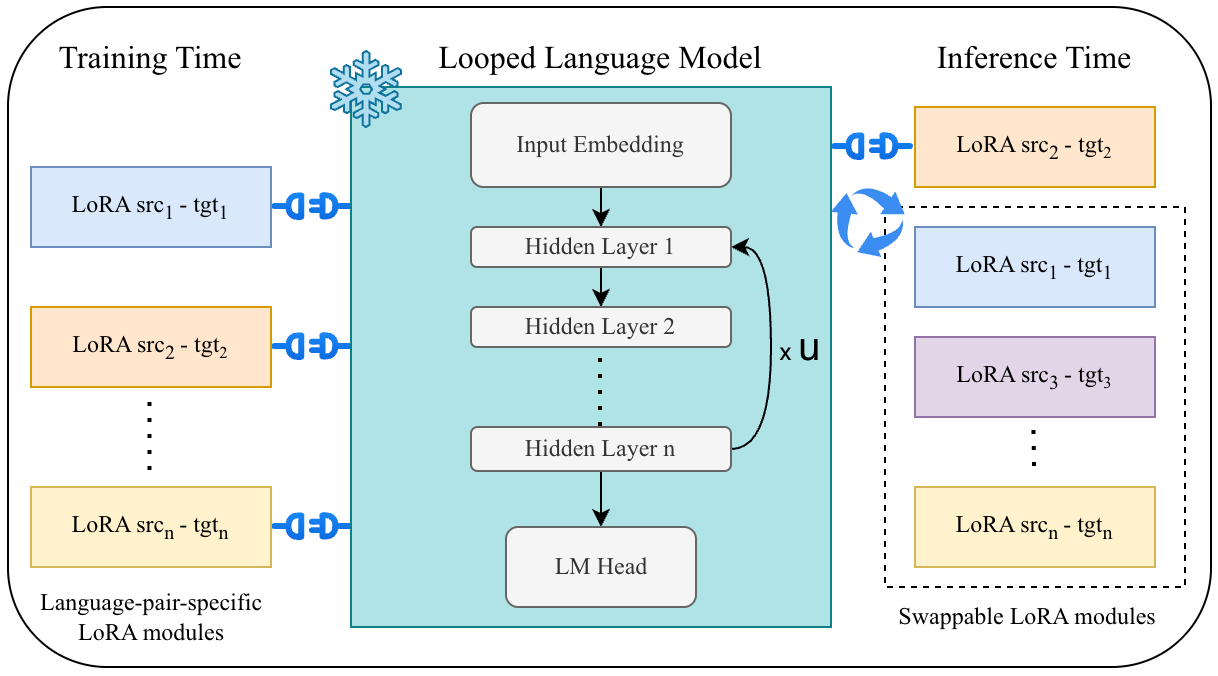}
	\caption{Overview of LatentMT.
	A frozen LoopLM backbone is paired with compact per-language-pair LoRA adapters, avoiding full-model duplication.
	At inference, the shared backbone is unrolled for recurrent latent reasoning before each next-token prediction, adding internal computation without adding parameters.
	Swappable adapters let one deployed backbone serve many translation directions.}
	\label{fig:overview}
\end{figure}

Latent reasoning is a paradigm in which a language model performs reasoning through continuous hidden states rather than externalizing reasoning steps as discrete text tokens~\cite{zhu2025scalinglatentreasoninglooped, geiping2025scalingtesttimecomputelatent,saunshi2025reasoning}.
Looped language models (LoopLMs) instantiate this paradigm by repeatedly applying shared Transformer weights to update these states before each next-token prediction~\cite{saunshi2025reasoning, gatmiry2024can, jeddi2026loopformer}.
This makes latent reasoning depth an architectural choice: the model can allocate more internal computation through additional recurrent steps without adding parameters at each step.
For machine translation (MT), the effectiveness of latent-reasoning LoopLMs and the relationship between recurrent depth and translation quality remain largely unexplored.

This shared-parameter approach contrasts with conventional non-loop LMs, which usually obtain more capacity by increasing model size~\cite{kaplan2020scalinglawsneurallanguage}. For MT, this alternative scaling strategy has important practical implications.
If latent reasoning can shift some of the burden from model size to repeated computation over a compact backbone, then strong translation systems become more plausible in resource-constrained settings, including edge devices and local deployments~\cite{lin2023mobilenmtenablingtranslation15mb}.
Such deployments can reduce dependence on cloud APIs, keep sensitive text closer to the user, and make MT more accessible to communities with limited or unreliable Internet access~\cite{xu2020edgeintelligencearchitectureschallenges}.

In this work, we introduce LatentMT, an MT approach built from compact latent-reasoning LoopLMs.
Figure~\ref{fig:overview} illustrates the approach: a single frozen LoopLM backbone applies its shared weights over several recurrent steps, and each language pair is served by a lightweight LoRA module that can be swapped in at inference time without touching the backbone.
We make recurrent depth an experimental variable and ask whether additional latent refinement steps improve translation quality.
Our study is organized around two research questions:
\begin{quote}
\begin{itemize}
    \item[\textbf{RQ1.}] Can compact latent-reasoning LoopLMs provide a competitive architecture for machine translation compared with larger regular language models?
    \item[\textbf{RQ2.}] Does increasing the number of latent reasoning steps improve MT performance, and where does this improvement saturate?
\end{itemize}
\end{quote}

We apply LatentMT to 32 language pairs spanning high-, medium-, and low-resource regimes, using a compact LoRA module~\cite{hu2022lora} for each pair over the same frozen backbone.
We compare against specialized MT systems, state-of-the-art models, and larger regular language models, including a Qwen3-8B baseline trained under the same conditions.
Despite relying on a 2.6B-parameter backbone that is three to five times smaller than leading comparison models, LatentMT achieves comparable performance overall and state-of-the-art results in some settings.
Translation quality improves with additional recurrent steps before saturating, and our analysis shows that most output-distribution refinement occurs in early recurrent transitions.

This work makes three contributions.
First, we introduce LatentMT and, to our knowledge, provide the first systematic study of latent reasoning for MT.
Second, we empirically compare LatentMT with explicit-reasoning regular language model, showing that compact latent-reasoning LoopLMs can provide a competitive architecture for MT.
Third, we study recurrent depth directly and show, supported by mechanistic and linguistic error analysis, that additional latent reasoning steps improve translation performance, but the gains saturate quickly after the first few steps.

%% file: section/related.tex
\section{Related Work}

\subsection{Machine Translation and Explicit Reasoning}

Chain-of-thought (CoT) prompting~\cite{NEURIPS2022_9d560961} has become a standard way to elicit explicit reasoning from large language models, especially on tasks where reasoning steps are useful.
Recent MT work suggests that this explicit reasoning can be helpful for translation.
Large reasoning models have been argued to improve MT by supporting coherence and the identification of cultural intent~\cite{liu2025newtrendsmodernmachine,ataee2026chainofthoughtreasoningimprovescontextaware}.
Empirical studies also report CoT gains in literary translation~\cite{wang-etal-2025-drt}. At the same time, these gains are not uniform.
Some studies find that explicit reasoning has setting-dependent effects~\cite{nguyen-xu-2025-reasoning, zebaze2025compositionaltranslationnovelllmbased}.
Other work indicates that explicit reasoning can add a resource-intensive generation stage without directly improving the final translation~\cite{wu-etal-2025-please,rajaee2026unlockingreasoningcapabilitymachine}.

This landscape motivates LatentMT.
We treat explicit reasoning as a strong comparison point for MT and ask a different question: whether latent reasoning can provide translation gains when the intermediate computation is performed inside hidden states rather than emitted as chain-of-thought text.
We isolate this effect by varying a fixed number of recurrent steps while training the model to generate only the target translation.
This allows us to measure how translation quality changes as more computation is allocated to latent refinement.

\subsection{Latent Reasoning vs. Explicit Reasoning}

Explicit and latent reasoning scale test-time computation along different axes.
CoT scales along the output-token sequence: more computation means longer generated reasoning traces~\cite{snell2025scaling}.
Latent reasoning instead scales along recurrent depth, moving intermediate computation into continuous hidden states~\cite{lu2025latentchainofthoughtdecodingdepthrecurrent}.
LoopLMs do so by repeatedly applying a shared stack of hidden layers to the evolving sequence state~\cite{zhu2025scalinglatentreasoninglooped,geiping2025scalingtesttimecomputelatent}.
Explicit CoT therefore scales computation along the output-token sequence, whereas LoopLMs scale computation along recurrent depth without emitting intermediate reasoning tokens.

LatentMT focuses on this depth-recurrent form of latent reasoning in the specific setting of machine translation.
For MT, the key question is whether a translation model can use additional recurrent hidden-state updates to improve the final target-language output without producing intermediate reasoning traces or other non-translation tokens.
Whereas prior LoopLM work primarily studies general reasoning and STEM-oriented tasks~\cite{hao2025traininglargelanguagemodels,wei2025simcotsupervisedimplicitchainofthought,deng2026llmlatentreasoningchain}, we examine how the same recurrent-depth axis transfers to MT and treat recurrent depth as a controlled experimental variable.

\subsection{Edge Computing Translation}

Machine translation can often be used on sensitive text, including personal communication, medical information, and business content~\cite{kamocki-oregan-2016-privacy}.
Cloud-based MT systems can provide strong quality, but they also require users to transmit source text to remote servers and depend on network connectivity.
For many users and communities, especially in low-connectivity or resource-constrained settings, local translation is therefore not only an efficiency goal but also an access goal~\cite{10539577}.
This makes compact, deployable MT models an important complement to very large cloud MT systems.

LatentMT is motivated by this deployment setting.
LoopLMs trade repeated computation for model size, and LoRA extends the saving to multi-language support: each pair is a lightweight attachment to the same frozen backbone.
Together they point toward edge intelligence, where reduced connectivity dependence keeps private text local~\cite{chabal2023achievingprivacypreservingstateoftheartedge}.

%% file: section/experiments.tex
\section{Experiments}

\subsection{Formulation}
\label{subsec:formulation}

We formulate LatentMT as parameter-efficient supervised fine-tuning of the pre-trained looped language model, natively trained with latent reasoning~\cite{zhu2025scalinglatentreasoninglooped}.
For a source sentence $x=(x_1,\ldots,x_m)$ and target translation $y=(y_1,\ldots,y_n)$, the model receives a chat-style translation prompt and is trained to produce the target translation as the assistant response.
Let $p(x)$ denote this source-conditioned prompt.
Training uses the concatenated sequence
\begin{equation}
s = [p(x); y; \mathrm{eos}],
\end{equation}
but masks all prompt positions, so gradients are taken only on target-side assistant tokens.

Let $E$ be the embedding layer, $F_{\Theta}$ the shared recurrent Transformer stack, and $W_{\ell m}$ the language-model head.
Let $u$ denote the fixed number of recurrent computation steps applied before predicting each next token.
For a fixed $u$,
\begin{align}
h^{(0)} &= E(s), \\
h^{(k)} &= F_{\Theta}(h^{(k-1)}), \qquad k=1,\ldots,u, \\
\pi^{(u)}_t &= \operatorname{softmax}\!\left(W_{\ell m} h^{(u)}_t\right), \\
P_{\Theta}^{u}(s_{t+1}) &= \pi^{(u)}_t[s_{t+1}]
= P_{\Theta}(s_{t+1}\mid s_{< t+1};u) .
\end{align}
In our experiments, we deactivate Ouro's adaptive early-exit mechanism and force every generated token to use the same pre-set number of recurrent steps.
This fixed-$u$ setting is necessary for analysis: with adaptive early exit enabled, different tokens may terminate at different recurrent steps, making it difficult to attribute translation behavior to a specific computation depth.
We therefore sweep $u$ across runs to study how recurrent depth affects machine translation performance while removing adaptive early exit as a confounder.

During fine-tuning, the pretrained backbone parameters $\Theta_0$ are frozen and only the LoRA adapter parameters \cite{hu2022lora} are updated.
This keeps the recurrent latent-reasoning backbone fixed while adapting the model to the translation distribution with a much smaller number of trainable parameters.

Let $\Psi$ denote all trainable adapter parameters and $\mathcal{T}(s)$ the unmasked target-token positions.
The training objective is the target-side causal language-modeling loss
\begin{equation}
\min_{\Psi}\;
\frac{1}{|\mathcal{D}|}
\sum_{(x,y)\in\mathcal{D}}
\frac{1}{|\mathcal{T}(s)|}
\sum_{t\in\mathcal{T}(s)}
-\log P_{\Theta_0,\Psi}^{u}(s_t).
\end{equation}
Thus, LatentMT adapts translation behavior through a small set of LoRA updates while preserving the latent-reasoning mechanism of the backbone model.

\subsection{Experimental Setup}
\label{subsec:exp_setup}

\paragraph{Models.}
Our main backbone model is Ouro-2.6B-Thinking\footnote{\url{https://huggingface.co/ByteDance/Ouro-2.6B-Thinking}.}~\cite{zhu2025scalinglatentreasoninglooped}, a 2.6B-parameter LoopLM natively trained with recurrent latent computation.
Ouro-2.6B-Thinking is trained predominantly on English data, with Chinese language appearing minimally during pretraining and being entirely absent from mid-training and post-training stages.
For the recurrent-depth study, we evaluate $u\in\{1,2,\dots,6\}$.
In the results tables we write LatentMT$_u$ and Ouro$_u$ for a model evaluated at a fixed recurrent depth $u$.
For comparison, we also include Qwen3-8B\footnote{\url{https://huggingface.co/Qwen/Qwen3-8B}}~\cite{yang2025qwen3technicalreport} as a larger non-looped explicit-reasoning LM, trained under the same dataset splits and training/evaluation protocol.

\paragraph{Datasets.}
We train and evaluate on three translation benchmark families.
English--Chinese language DRT dataset\footnote{\url{https://github.com/krystalan/DRT}}~\cite{wang-etal-2025-drt} serves as a high-resource representative, while English--Egyptian Arabic ArzEn-MultiGenre dataset\footnote{\url{https://data.mendeley.com/datasets/6k97jty9xg/4}}~\cite{ALSABBAGH2024110271} serves as a mid-resource representative.
To include low-resource languages, we use OLDI/Flores, a randomly sampled subset of 30 languages in the intersection of OLDI-seed dataset\footnote{\url{https://huggingface.co/datasets/openlanguagedata/oldi_seed}}~\cite{seed-23} and Flores-200 benchmark\footnote{\url{https://github.com/facebookresearch/flores}}~\cite{nllbteam2022languageleftbehindscaling}, training on OLDI-seed and evaluating on Flores-200.
Table~\ref{tab:dataset_sizes} summarizes the train, validation, and test sizes for all three dataset families.

\input{tables/DatasetSizes}

\paragraph{Training.}
To enable lightweight training, we fine-tune with LoRA while keeping the pretrained backbone frozen.
For DRT and ArzEn-MultiGenre, LoRA uses rank 32, $\alpha=64$, and dropout 0.05.
For OLDI/Flores, which has a smaller training split, LoRA uses rank 16, $\alpha=32$, and dropout 0.05.
We train DRT for three epochs, ArzEn-MultiGenre for six epochs, and OLDI/Flores for four epochs.
All runs use AdamW, learning rate $2\times10^{-4}$, cosine scheduling, warmup ratio 0.03, maximum sequence length 3{,}072, per-device batch size eight, and gradient accumulation one.
Training masks prompt tokens and optimizes only assistant-side target tokens.
For each language pair, the training is conducted on single Nvidia A100 (40GB) or H100 (80GB) GPU.

\paragraph{Evaluation.}
At inference time, each model generates translations from the test split of each included dataset.
For LatentMT, generation uses the fixed number of recurrent step(s) $u$ specified for the run.
Decoding uses temperature 1.0, nucleus sampling with top-$p=0.7$.
We evaluate with zero-shot inference for trained models and few-shot inference for out-of-box models; few-shot prompts use five training exeamplares to perform in-context learning.

We report BLEU and chrF++ via SacreBLEU\footnote{\url{https://github.com/mjpost/sacrebleu}}
, together with the neural metrics COMET\footnote{
\url{https://huggingface.co/Unbabel/wmt22-comet-da}}~\cite{rei-etal-2022-comet} and COMETKiwi\footnote{\url{https://huggingface.co/Unbabel/wmt22-cometkiwi-da}}~\cite{rei-etal-2022-cometkiwi} using COMET framework\footnote{\url{https://github.com/Unbabel/COMET}}~\cite{stewart-etal-2020-comet}.
BLEU uses the \texttt{zh} tokenizer, which specifically supports Chinese text, for DRT; and the \texttt{flores200} tokenizer, which supports all the languages covered, for ArzEn-MultiGenre and OLDI/Flores.
For ArzEn-MultiGenre, we report an overall BLEU calculated over the combined test sets (304, 518, and 844, respectively) of all three genres (Songs, Novels, Subtitles)\footnote{The original ArzEn-MultiGenre dataset paper~\cite{ALSABBAGH2024110271} reports split \emph{sizes} but does not release the actual train/validation/test partition.
We therefore randomly sample validation and test splits matching the reported sizes (1{,}666 segments each; per genre 304/518/844 for Songs/Novels/Subtitles) and assign all remaining examples to the training split.}.

%% file: tables/DatasetSizes.tex
\begin{table}[t]
\centering
\footnotesize
\begin{tabular}{@{}lrrr@{}}
\toprule
Dataset & Train & Validation & Test \\
\midrule
DRT & 19{,}264 & 1{,}000 & 2{,}000 \\
ArzEn-MultiGenre & 22{,}224 & 1{,}666 & 1{,}666 \\
OLDI/Flores & 6{,}193 & 997 & 1{,}012 \\
\bottomrule
\end{tabular}
\caption{Train, validation, and test sizes for the three dataset families. OLDI/Flores includes 30 target languages; sizes are per target language.}
\label{tab:dataset_sizes}
\end{table}

%% file: section/results.tex
\section{Results}

\subsection{High-resource DRT}

\input{tables/DRT}

Table~\ref{tab:drt} reports results on DRT.
LatentMT produces significant improvements at every recurrent depth, indicating that the framework can further improve translation quality even when the out-of-box backbone already shows decent performance at larger recurrent depths.
Out-of-box Ouro remains at 0.00--9.02 BLEU, 32.01--57.27 COMET, and 21.38--50.23 COMETKiwi across the sweep, whereas LatentMT reaches 20.25--29.37 BLEU, 66.06--76.57 COMET, and 58.23--69.21 COMETKiwi.
Most of the gain from additional recurrent computation appears early: moving from $u=1$ to $u=2$ adds 7.65 BLEU, 9.35 COMET, and 9.86 COMETKiwi, after which the curve changes only marginally through $u=6$.

Against Qwen3-8B SFT, our same-protocol control, the best LatentMT checkpoint is higher on all three metrics despite using a substantially smaller 2.6B-parameter backbone.
This comparison is especially notable because Ouro has only minimal Chinese language exposure during pretraining and no Chinese language data in mid-training or post-training, whereas the comparison systems are multilingual or DRT-specific models with substantially stronger Chinese language supervision.
Compared with the larger, domain-specialized reference systems, LatentMT exhibits strong parameter efficiency.
DRT-14B and ExTrans-7B retain a slight edge on selected metrics, but the best LatentMT-2.6B checkpoint closes much of the performance gap while operating at a 3$\times$ to 5$\times$ parameter disadvantage.
Thus, in the high-resource setting, LatentMT clearly improves over a larger Chinese-language-capable LLM trained under the same conditions and approaches the performance of systems that are substantially larger and specialized for literary translation, the domain the DRT dataset focuses on.

\subsection{Mid-resource ArzEn-MultiGenre}

\input{tables/ArzEnMultiGenre}

Table~\ref{tab:arzenmultigenre} gives the overall BLEU results for ArzEn-MultiGenre, our mid-resource English-to-Egyptian-Arabic benchmark.
Unlike on DRT, out-of-box Ouro is effectively unable to translate this benchmark: overall BLEU stays between 0.00 and 0.20 across recurrent depths, with every genre-level score at or below 0.37.
Fine-tuning therefore turns a near-zero out-of-box system into a strong mid-resource translator, indicating that the LatentMT framework is effective even when the out-of-box backbone is weak at translating into the target language.
LatentMT achieves the best overall score among all compared systems, peaking at 24.47 BLEU with $u=4$.
This is 7.28 BLEU above fine-tuned Google NMT and 5.31 BLEU above Qwen3-8B SFT, indicating that the recurrent model remains competitive even against a much larger multilingual LM.

The recurrent-depth trend again shows rapid early improvement followed by saturation.
LatentMT rises from 15.35 at $u=1$ to 21.29 at $u=2$ and 23.30 at $u=3$, then fluctuates within a narrow band from $u=4$ to $u=6$.
At the genre level, LatentMT shows robust cross-domain adaptability rather than uniform dominance.
Google NMT fine-tuned on this benchmark's own data remains strongest on the more stylistic Songs and Novels genres, but LatentMT is substantially stronger on the highly conversational Subtitles domain, outperforming fine-tuned Google NMT by 17.93 BLEU points (36.98 vs. 19.05).
This large Subtitle gain, together with competitive performance on the other weighted genres, drives LatentMT to the highest overall weighted BLEU among all compared systems.
This suggests that LatentMT provides a robust mid-resource translation model, even though the backbone has limited capabilities before training.

\subsection{Low-resource OLDI/Flores}

\input{tables/Flores200_30langs}

Table~\ref{tab:flores200_30langs} reports BLEU and chrF++ on the 30 displayed low-resource OLDI/Flores target language/script configurations.
LatentMT obtains the highest mean BLEU and chrF++ among the displayed systems, reaching 12.71 BLEU and 29.94 chrF++.
It is slightly ahead of EMMA-500-Llama3-8B on both averages (12.51 BLEU and 29.13 chrF++) and clearly ahead of Qwen3-8B SFT (8.27 BLEU and 21.76 chrF++).
LatentMT is able to achieve the best average score on both reported metrics despite using a much smaller 2.6B-parameter backbone.

Overall, the three benchmarks show that the LatentMT framework yields strong translation gains across high-, mid-, and low-resource settings.
They also reveal a consistent recurrent-depth pattern: most gains appear when moving from low to moderate fixed recurrent depths, while higher depths mainly preserve or lightly refine quality.
The next section examines this saturation behavior directly and includes a manual linguistic analysis to further support the effectiveness of the LatentMT framework beyond automatic metrics.

%% file: tables/DRT.tex
\begin{table}[t]
\centering
\resizebox{\columnwidth}{!}{
\begin{tabular}{l r S[table-format=2.2,detect-all] S[table-format=2.2,detect-all] S[table-format=2.2,detect-all]}
\toprule
Model & Size & {BLEU} & {COMET} & {COMETKiwi} \\
\midrule
\multicolumn{5}{l}{\textit{Reference systems (larger, literary-specialized)}} \\
DRT-8B & 8B & 32.67 & 78.81 & 70.85 \\
DRT-14B & 14B & {\underline{36.46}} & {\underline{80.64}} & 72.11 \\
DeepTrans-7B & 7B & {21.55$^{\star}$} & {78.63$^{\star}$} & {71.82 / 72.45$^{\star}$} \\
ExTrans-7B & 7B & {16.32$^{\star}$} & {62.92$^{\star}$} & {\underline{74.23} / 54.93$^{\star}$} \\
\midrule
\multicolumn{5}{l}{\textit{Backbone (untrained, out-of-box)}} \\
Ouro$_{1}$ & 2.6B & 0.00 & 32.01 & 21.38 \\
Ouro$_{2}$ & 2.6B & 0.00 & 33.74 & 22.85 \\
Ouro$_{3}$ & 2.6B & 1.21 & 42.69 & 32.85 \\
Ouro$_{4}$ & 2.6B & 6.07 & 52.53 & 44.17 \\
Ouro$_{5}$ & 2.6B & 9.02 & 57.27 & 50.23 \\
Ouro$_{6}$ & 2.6B & 8.87 & 57.26 & 49.99 \\
\midrule
\multicolumn{5}{l}{\textit{Same-protocol systems (ours)}} \\
Qwen3-8B SFT & 8B & 13.19 & 54.82 & 45.14 \\
LatentMT$_{1}$ & 2.6B & 20.25 & 66.06 & 58.23 \\
LatentMT$_{2}$ & 2.6B & 27.90 & 75.41 & 68.09 \\
LatentMT$_{3}$ & 2.6B & 29.18 & 76.32 & 68.80 \\
LatentMT$_{4}$ & 2.6B & 29.34 & 76.51 & 69.01 \\
LatentMT$_{5}$ & 2.6B & \textbf{29.37} & \textbf{76.57} & \textbf{69.21} \\
LatentMT$_{6}$ & 2.6B & 29.14 & 76.36 & 68.84 \\
\bottomrule
\end{tabular}
}
\caption{DRT results (English-to-Chinese). LatentMT and its out-of-box Ouro backbone use a 2.6B-parameter model, while DRT~\cite{wang-etal-2025-drt}, DeepTrans~\cite{wang2025deeptransdeepreasoningtranslation}, and ExTrans~\cite{wang2025extransmultilingualdeepreasoning} are state-of-the-art 7B--14B systems built specifically for literary translation, the domain the DRT dataset focuses on. Among systems trained under our identical protocol, \textbf{bold} marks the best value per metric; \underline{underline} marks the best value across all systems. Subscripts denote the fixed number of recurrent steps $u$. Superscript stars mark reference values we reproduced by evaluating those models ourselves; unstarred reference values are from the original papers.}
\label{tab:drt}
\end{table}

%% file: tables/ArzEnMultiGenre.tex
\begin{table}[t]
\centering
\setlength{\aboverulesep}{0pt}
\setlength{\belowrulesep}{0pt}
\renewcommand{\arraystretch}{1.2}
\resizebox{\columnwidth}{!}{
\begin{tabular}{l r r r r |r}
\toprule
Model & Size & Songs & Subtitles & Novels & Overall \\
\midrule
\multicolumn{5}{l|}{\textit{Reference systems (commercial and domain-adapted)}} & \\
Google NMT & Unk & 8.95 & 12.54 & 11.84 & 11.67 \\
Fine-tuned Google NMT & Unk & \underline{11.87} & 19.05 & \underline{17.28} & 17.19 \\
\midrule
\multicolumn{5}{l|}{\textit{Backbone (untrained, out-of-box)}} & \\
Ouro$_{1}$ & 2.6B & 0.01 & 0.01 & 0.00 & 0.00 \\
Ouro$_{2}$ & 2.6B & 0.00 & 0.05 & 0.00 & 0.02 \\
Ouro$_{3}$ & 2.6B & 0.01 & 0.19 & 0.00 & 0.10 \\
Ouro$_{4}$ & 2.6B & 0.01 & 0.37 & 0.02 & 0.20 \\
Ouro$_{5}$ & 2.6B & 0.06 & 0.32 & 0.01 & 0.18 \\
Ouro$_{6}$ & 2.6B & 0.03 & 0.31 & 0.04 & 0.18 \\
\midrule
\multicolumn{5}{l|}{\textit{Same-protocol systems (ours)}} & \\
Qwen3-8B SFT & 8B & 4.28 & 30.48 & 9.45 & 19.16 \\
LatentMT$_{1}$ & 2.6B & 4.53 & 23.11 & 9.05 & 15.35 \\
LatentMT$_{2}$ & 2.6B & 6.49 & 31.81 & 12.84 & 21.29 \\
LatentMT$_{3}$ & 2.6B & 6.39 & 35.20 & 13.85 & 23.30 \\
LatentMT$_{4}$ & 2.6B & \textbf{7.61} & \underline{\textbf{36.98}} & 13.96 & \underline{\textbf{24.47}} \\
LatentMT$_{5}$ & 2.6B & 6.56 & 36.60 & 13.41 & 23.91 \\
LatentMT$_{6}$ & 2.6B & 7.14 & 36.04 & \textbf{14.80} & 24.16 \\
\bottomrule
\end{tabular}
}
\caption{BLEU on ArzEn-MultiGenre (English-to-Egyptian Arabic). LatentMT and its out-of-box Ouro backbone use a 2.6B-parameter model, while the reference systems are a commercial NMT engine and a variant fine-tuned on this benchmark's own data~\cite{ALSABBAGH2024110271}. Overall is weighted by the Songs/Subtitles/Novels test sizes (304/844/518). Among systems trained under our identical protocol, \textbf{bold} marks the best value per metric; \underline{underline} marks the best value across all systems. Subscripts denote the fixed number of recurrent steps $u$.}
\label{tab:arzenmultigenre}
\end{table}

%% file: tables/Flores200_30langs.tex
\begin{table}[t]
\centering
\scriptsize
\resizebox{\linewidth}{!}{
\begin{tabular}{ll rr rr rr}
\toprule
 &  & \multicolumn{2}{c}{LatentMT-2.6B$_4$} & \multicolumn{2}{c}{Qwen3-8B SFT} & \multicolumn{2}{c}{EMMA-500-8B} \\
\cmidrule(lr){3-4} \cmidrule(lr){5-6} \cmidrule(lr){7-8}
Target & Lang Family & BLEU & chrf++ & BLEU & chrf++ & BLEU & chrf++ \\
\midrule
ace\_Arab & Austronesian & \underline{\textbf{8.52}} & \underline{\textbf{21.15}} & 4.49 & 16.25 & 1.72 & 12.13 \\
ary\_Arab & Afro-Asiatic & \textbf{13.25} & \textbf{30.89} & 9.47 & 22.75 & \underline{13.67} & \underline{32.70} \\
bam\_Latn & Mande & \underline{\textbf{6.03}} & \underline{\textbf{24.96}} & 2.23 & 13.81 & 3.15 & 20.85 \\
ban\_Latn & Austronesian & \textbf{12.61} & \textbf{37.98} & 9.91 & 31.86 & \underline{16.93} & \underline{42.58} \\
bjn\_Arab & Austronesian & \underline{\textbf{12.05}} & \underline{\textbf{25.31}} & 5.44 & 14.65 & 2.59 & 15.68 \\
bjn\_Latn & Austronesian & \textbf{14.24} & \textbf{41.24} & 7.69 & 25.45 & \underline{24.03} & \underline{50.05} \\
bug\_Latn & Austronesian & \underline{\textbf{6.51}} & \underline{\textbf{29.90}} & 4.40 & 22.50 & 1.51 & 17.47 \\
crh\_Latn & Turkic & \textbf{14.04} & \textbf{34.85} & 5.50 & 18.84 & \underline{16.97} & \underline{38.15} \\
dik\_Latn & Nilotic & \underline{\textbf{5.61}} & \underline{\textbf{22.80}} & 3.69 & 14.80 & 1.39 & 12.65 \\
dzo\_Tibt & Sino-Tibetan & 2.32 & \textbf{15.96} & \textbf{2.55} & 14.58 & \underline{4.35} & \underline{26.61} \\
fur\_Latn & Indo-European & \textbf{29.53} & \textbf{48.85} & 15.17 & 28.05 & \underline{34.17} & \underline{53.00} \\
fuv\_Latn & Atlantic-Congo & \underline{\textbf{3.62}} & \underline{\textbf{20.46}} & 2.78 & 16.13 & 0.81 & 11.36 \\
kas\_Arab & Indo-European & \underline{\textbf{12.56}} & \textbf{28.22} & 5.81 & 15.40 & 11.83 & \underline{28.94} \\
kas\_Deva & Indo-European & \underline{\textbf{3.28}} & \underline{\textbf{15.47}} & 2.99 & 13.86 & 2.07 & 13.93 \\
knc\_Arab & Saharan & \underline{\textbf{12.48}} & \underline{\textbf{16.20}} & 11.28 & 14.09 & 0.52 & 3.70 \\
knc\_Latn & Saharan & \underline{\textbf{5.00}} & \underline{\textbf{23.39}} & 2.68 & 16.76 & 0.98 & 13.30 \\
lij\_Latn & Indo-European & \underline{\textbf{29.53}} & \textbf{47.25} & 19.50 & 34.13 & 28.80 & \underline{47.52} \\
lim\_Latn & Indo-European & \textbf{19.20} & \textbf{40.09} & 14.49 & 31.67 & \underline{22.69} & \underline{45.05} \\
lmo\_Latn & Indo-European & \textbf{9.97} & \textbf{31.79} & 7.09 & 24.51 & \underline{11.39} & \underline{33.33} \\
ltg\_Latn & Indo-European & \textbf{25.24} & \textbf{44.27} & 20.74 & 37.27 & \underline{26.62} & \underline{46.66} \\
mni\_Beng & Sino-Tibetan & \textbf{16.50} & \textbf{28.12} & 14.96 & 25.83 & \underline{21.93} & \underline{35.10} \\
nus\_Latn & Nilotic & \underline{\textbf{10.39}} & \underline{\textbf{26.22}} & 7.09 & 19.40 & 0.80 & 9.47 \\
pbt\_Arab & Indo-European & \textbf{13.74} & \textbf{31.04} & 6.61 & 17.72 & \underline{16.75} & \underline{34.60} \\
prs\_Arab & Indo-European & \textbf{15.06} & \textbf{36.32} & 10.81 & 26.57 & \underline{20.97} & \underline{42.58} \\
shn\_Mymr & Tai-Kadai & 0.09 & 7.24 & \textbf{2.57} & \textbf{15.64} & \underline{6.00} & \underline{25.89} \\
srd\_Latn & Indo-European & \textbf{25.88} & \textbf{47.50} & 16.13 & 34.17 & \underline{35.17} & \underline{55.19} \\
szl\_Latn & Indo-European & \textbf{23.49} & \textbf{41.13} & 12.26 & 25.82 & \underline{26.17} & \underline{44.57} \\
taq\_Latn & Afro-Asiatic & \underline{\textbf{3.78}} & \underline{\textbf{21.19}} & 2.31 & 15.78 & 0.23 & 6.07 \\
taq\_Tfng & Afro-Asiatic & 3.06 & 11.83 & \underline{\textbf{4.23}} & \underline{\textbf{14.54}} & 0.60 & 8.99 \\
vec\_Latn & Indo-European & \underline{\textbf{23.83}} & \underline{\textbf{46.57}} & 13.16 & 29.91 & 20.60 & 45.80 \\
\midrule
Average &  & \underline{\textbf{12.71}} & \underline{\textbf{29.94}} & 8.27 & 21.76 & 12.51 & 29.13 \\
\bottomrule
\end{tabular}
}
\caption{BLEU and chrf++ on 30 language pairs of Flores200 (English is always the source language). LatentMT-2.6B$_4$ and Qwen3-8B SFT are trained under our identical protocol; EMMA-500-8B refers to the state-of-the-art EMMA-500-Llama3-8B model~\cite{ji2025emma2}. Between the two same-protocol systems, \textbf{bold} marks the better value per metric; \underline{underline} marks the best value across all three systems. LatentMT-2.6B$_4$ uses a fixed recurrent depth of $u=4$.}
\label{tab:flores200_30langs}
\end{table}

%% file: section/analysis2.tex
\section{Analysis}
\label{sec:analysis}

\subsection{Performance Scaling with Recurrent Steps}
\label{subsec:recurrent-step-scaling}

We ask how performance scales with the number of recurrent steps.
Since each additional recurrent step reapplies the same shared Transformer stack before producing the next-token distribution, increasing $u$ gives the model more latent refinement compute without changing the parameter count.
The relevant question is therefore not whether more steps can add compute, but whether that compute continues to translate into better MT quality.

\begin{figure}[t]
	\centering
	\includegraphics[width=\linewidth]{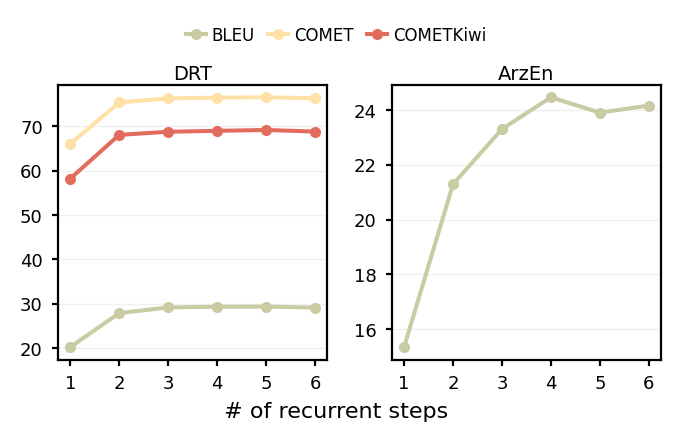}
	\caption{Performance scaling with the fixed number of recurrent steps.
	DRT reports BLEU, COMET, and COMETKiwi, while ArzEn-MultiGenre reports BLEU.
	Most gains occur when moving from low to moderate fixed recurrent depths, followed by saturation at higher depths.}
	\label{fig:recurrent-step-scaling}
\end{figure}

Figure~\ref{fig:recurrent-step-scaling} shows a consistent early-gain pattern across the two datasets.
On DRT, the dominant improvement occurs between $u=1$ and $u=2$.
Subsequent depths mainly produce small adjustments: the best BLEU and COMET scores occur at $u=5$, but the gains from $u=2$ to $u=5$ are only 1.47 BLEU, 1.16 COMET, and 1.12 COMETKiwi.
ArzEn-MultiGenre shows a slightly longer rise, improving from 15.35 BLEU at $u=1$ to 24.47 at $u=4$, then remaining close to that level through $u=6$.
Together, these trends suggest that recurrent latent computation is useful for MT, but its marginal returns concentrate in the first few steps.

We further validate this trend with a human linguistic error analysis of DRT and ArzEn-MultiGenre outputs at every recurrent depth.
For each dataset, translation errors were annotated by a trained linguist who is also a native speaker of the dataset language, using the LQM taxonomy~\cite{magdy-etal-2026-lqm}.
The error counts mirror the automatic metrics: errors drop sharply from low $u$ and then flatten, with severe errors concentrated at low recurrent depths.
Thus, the early-gain-then-saturation pattern is supported by both automatic metrics and expert assessment; the full analysis appears in Appendix~\ref{sec:appendix-lqm}.

The same saturation pattern appears inside the recurrent computation.
For a generated token, let $h_k \in \mathbb{R}^{d}$ be the hidden representation of the last hidden layer, the layer immediately preceding the language-model head, produced at recurrent step $k$.
For adjacent steps $k\rightarrow k+1$ we report the cosine distance
\begin{equation}
d_{\cos}(k) = 1 - \frac{h_k^{\top} h_{k+1}}{\lVert h_k\rVert\,\lVert h_{k+1}\rVert},
\label{eq:cosine-distance}
\end{equation}
Let $t_k = \arg\max\!\left(W_{\ell m} h_k\right)$ be the top-1 token at step $k$.
We also report the top-1 flip indicator
\begin{equation}
\mathrm{flip}(k) = \mathds{1}\!\left[\,t_k \neq t_{k+1}\,\right].
\label{eq:top1-flip}
\end{equation}

\begin{figure}[t]
	\centering
	\includegraphics[width=\columnwidth]{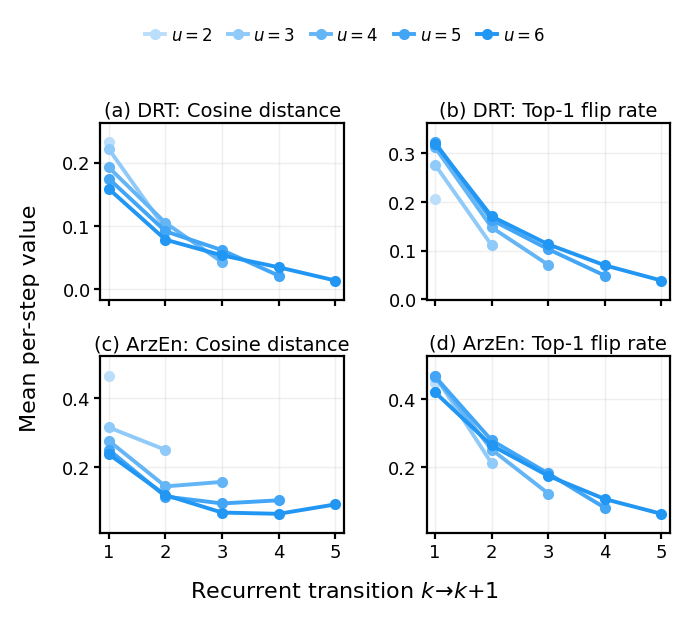}
	\caption{Mechanistic saturation across recurrent transitions.
	The columns show cosine distance and top-1 flip rate for DRT (top) and ArzEn-MultiGenre (bottom).
	Later transitions show much smaller changes in both metrics, indicating that recurrent updates diminish as depth increases.
	This mechanistic flattening aligns with the performance saturation at larger recurrent depths.}
	\label{fig:recurrent-step-mechanistic-saturation}
\end{figure}

Figure~\ref{fig:recurrent-step-mechanistic-saturation} plots, for each transition, the mean of these two metrics over generated tokens and examples.
Let $M\in\{d_{\cos},\mathrm{flip}\}$ be either metric, $\mathcal{E}$ be index examples, and $\mathcal{T}_e$ be the generated token positions of example $e$.
The reported per-transition mean at recurrent depth $u$ is
\begin{equation}
\overline{M}_u(k) = \frac{1}{|\mathcal{E}|}\sum_{e\in\mathcal{E}} \frac{1}{|\mathcal{T}_e|}\sum_{i\in\mathcal{T}_e} M(k;e,i).
\label{eq:per-transition-mean}
\end{equation}
Both metrics decline across successive transitions: the change is largest at the first transition and shrinks toward zero at the last.
At $u=6$ on DRT, for instance, the mean cosine distance falls from 0.159 at the first transition to 0.014 at the last, and the mean top-1 flip rate from 0.320 to 0.039; ArzEn-MultiGenre shows the same contraction.
The full per-transition means for every recurrent-depth checkpoint are reported in Table~\ref{tab:ut-transition-means} (Appendix~\ref{sec:appendix-transitions}).
By the final transitions the top-1 token rarely changes, indicating that the token preference has largely settled.
This stabilized token preference helps explain why translation metrics saturate at larger $u$.
Taken together, the performance scaling in Figure~\ref{fig:recurrent-step-scaling}, the mechanistic saturation in Figure~\ref{fig:recurrent-step-mechanistic-saturation}, and the linguistic error analysis in Appendix~\ref{sec:appendix-lqm} align with and support the same finding: recurrent computation improves translation quickly at lower $u$ and then saturates at larger $u$.

\subsection{Efficiency Analysis}
\label{subsec:efficiency}

\begin{figure*}[t]
	\centering
	\includegraphics[width=\textwidth]{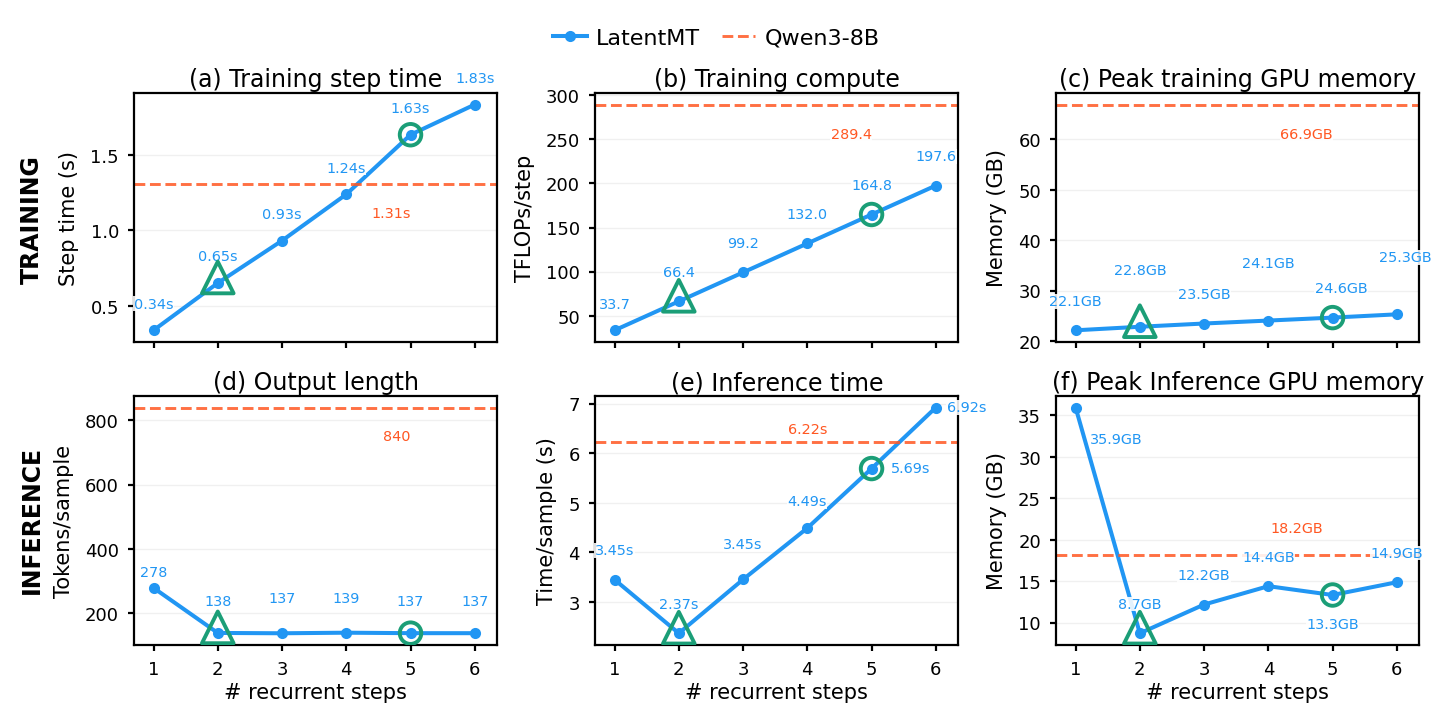}
	\caption{Efficiency comparison of LatentMT against Qwen3-8B on DRT.
	Scaling computation through recurrent depth, LatentMT keeps both training and inference GPU memory well below Qwen3-8B while also achieving lower end-to-end inference time.
	The circled marker ($u=5$) denotes the quality-optimal recurrent depth and the triangle marker ($u=2$) the most cost-efficient one.}
	\label{fig:efficiency}
\end{figure*}

We ask whether LatentMT, a LoopLM backbone with an attached LoRA module, offers a better efficiency--quality tradeoff than a conventional language model, comparing it against a Qwen3-8B reasoning model trained under identical conditions on DRT.
Our analysis centers on two operating points: the quality-optimal depth $u=5$ and a cost-efficient depth $u=2$.
The latter is chosen by knee-point detection\footnote{\url{https://github.com/arvkevi/kneed}}~\cite{satopa2011kneedle} on the average of BLEU, COMET, and COMETKiwi against each of four efficiency axes (training step time, peak training memory, inference time per sample, and peak inference memory), which unanimously place the knee at $u=2$ (triangle in Figure~\ref{fig:efficiency}).
Because quality saturates early, $u=2$ retains 97.9\% of the peak quality reached at $u=5$, making it a near-tied but far more resource-efficient operating point.

Weight sharing keeps LatentMT's GPU memory footprint small: each recurrent step reapplies the same Transformer stack rather than duplicating parameters, so both training and inference peak memory grow only slowly with $u$ (panels~(c) and~(f)) and stay well below Qwen3-8B.
Both operating points inherit this advantage, cutting peak training memory by over 63\% and training compute by 43\% ($u=5$) to 77\% ($u=2$) relative to Qwen3-8B; the cost-efficient $u=2$ additionally halves the per-step training time.

The inference behavior of the two models differs sharply.
Because the reasoning baseline emits a long explicit chain of thought before its answer whereas LatentMT refines in latent recurrent steps and emits only the translation, LatentMT produces significantly shorter outputs (panel~(d)).
This brevity keeps LatentMT below Qwen3-8B on both inference axes: it lowers end-to-end latency by 8.5\% at $u=5$ and 61.9\% at the more cost-efficient $u=2$, and peak inference memory by 26.9\% and 52.2\% respectively.

Overall, LatentMT is more efficient than the reasoning baseline at both operating points, with recurrent depth serving as a controllable dial between the cost-efficient $u=2$ and the quality-optimal $u=5$.

%% file: section/conclusion.tex
\section{Conclusion}

We presented \textbf{LatentMT}, a compact MT framework that adapts latent-reasoning looped language models to translation with a frozen 2.6B-parameter recurrent backbone and lightweight LoRA modules.
Across 32 translation directions, LatentMT remains competitive in high-resource Chinese and achieves state-of-the-art results in both Egyptian Arabic and low-resource languages, despite being three to five times smaller than several leading comparison models.
Our analysis further shows that recurrent latent computation improves translation most strongly at early depths, then saturates as hidden representations and token preferences stabilize across later steps.
Together with the efficiency results, these findings show that LatentMT offers a strong quality--compute tradeoff and an effective alternative to MT systems built upon regular language models.

%% file: section/appendix.tex
\onecolumn
\section{Linguistic Error Analysis}
\label{sec:appendix-lqm}

\begin{figure}[h]
	\centering
	\includegraphics[width=\textwidth]{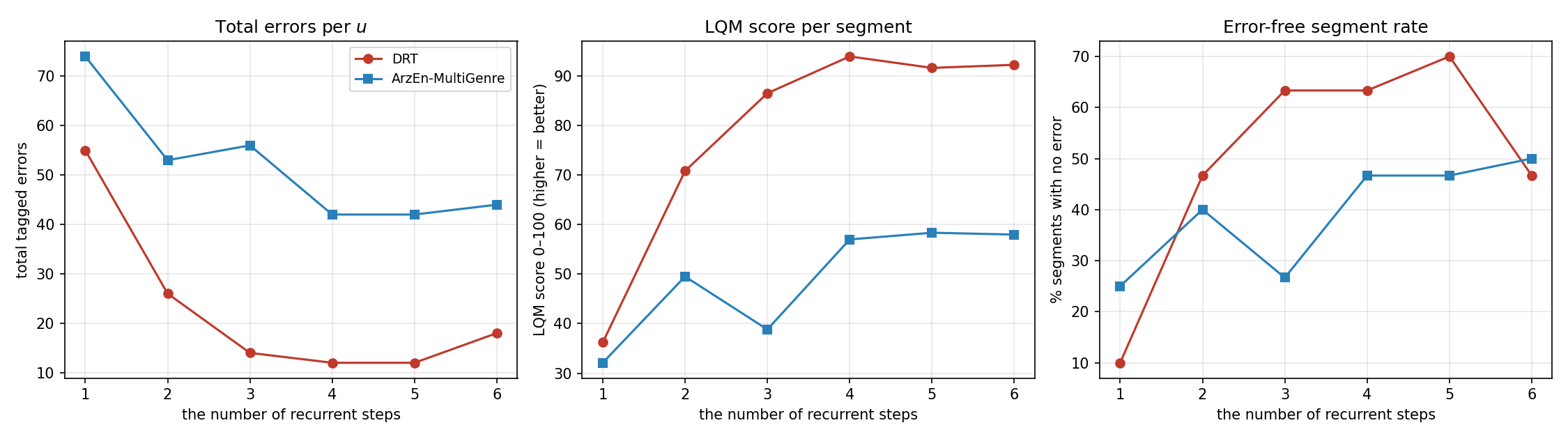}
	\caption{Human linguistic error analysis across recurrent depths for DRT and ArzEn-MultiGenre.
	Each panel reports, against the number of recurrent steps $u$, the total tagged errors, the mean LQM quality score per segment (0--100, higher is better), and the error-free segment rate.
	Both datasets improve steeply over the first few steps and then saturate.}
	\label{fig:appendix-lqm-overview}
\end{figure}

To demonstrate the recurrent-step scaling trends of Section~\ref{subsec:recurrent-step-scaling} with human assessment, we conducted a manual linguistic error analysis of the model outputs at every recurrent depth.
For each of DRT (30 segments) and ArzEn-MultiGenre (60 segments, 20 per genre), the six recurrent-depth outputs ($u=1$ through $u=6$) were produced for the same source segments and annotated by a trained linguist who is a native speaker of the relevant language.
Following the LQM framework~\cite{magdy-etal-2026-lqm}, the annotator labeled each translation error with a fine-grained error type and a severity (Minor, Major, or Critical).
We summarize each recurrent depth with the total number of tagged errors, the mean LQM quality score per segment, and the fraction of segments with no tagged errors, as shown in Figure~\ref{fig:appendix-lqm-overview}.

The analysis reproduces the early-gain-then-saturation pattern observed under automatic metrics.
Figure~\ref{fig:appendix-lqm-overview} shows that both datasets improve sharply as the fixed recurrent-depth setting increases from low values and then level off.
On DRT, mean errors per segment fall from 1.83 at $u=1$ to 0.47 by $u=3$ and remain low thereafter, while the corresponding LQM score rises from 36.2 at $u=1$ to 86.5 at $u=3$ and exceeds 90 from $u=4$ onward.
ArzEn-MultiGenre follows the similar trajectory: errors per segment decline from 1.23 at $u=1$ to 0.70 by $u=4$,  while the LQM score increases overall from 32.1 to about 58 and the error-free segment rate doubles from 25.0\% to 50.0\%; the largest gains occur as the fixed recurrent-depth setting increases up to $u=4$, with little movement thereafter.

\begin{figure}[h]
	\centering
	\includegraphics[width=0.8\textwidth]{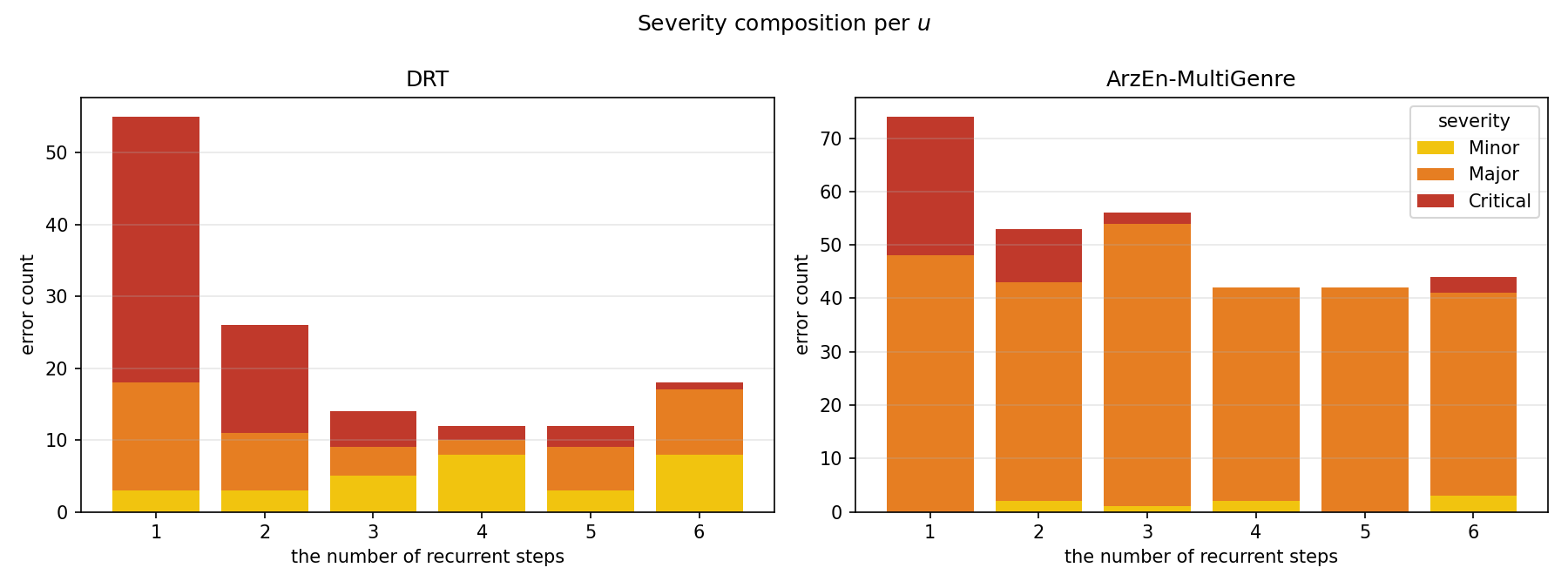}
	\caption{Severity composition of tagged errors across recurrent depths, for DRT (left) and ArzEn-MultiGenre (right).
	Critical errors are concentrated at low $u$ and become much rarer at higher fixed-$u$ settings, leaving mostly Minor and Major residual errors.}
	\label{fig:appendix-lqm-severity}
\end{figure}

The severity breakdown in Figure~\ref{fig:appendix-lqm-severity} explains why quality improves faster than the raw error count suggests: low-$u$ settings contain many of the most damaging errors, whereas higher fixed-$u$ settings contain far fewer of them.
Table~\ref{tab:lqm-low-ut-error-examples} gives representative examples in which annotated errors present at low $u$ are absent when the same source segment is translated with a higher fixed-$u$ setting.
\input{tables/lqm_low_ut_error_examples}
Critical errors are concentrated in the low-$u$ outputs and become much less common at higher $u$: on DRT they fall from 37 at $u=1$ to 5 by $u=3$ and 1 by $u=6$, while on ArzEn-MultiGenre they fall from 26 at $u=1$ to 2 by $u=3$, disappear at $u=4$--$5$, and reappear with small amount at $u=6$.
Taken together, the linguistic analysis shows that the behavior identified by our automatic evaluation is also supported at the level of concrete, human-verified translation errors, strengthening the robustness of our conclusions.

\section{Per-Transition Mechanistic Means}
\label{sec:appendix-transitions}

Table~\ref{tab:ut-transition-means} includes the full per-transition
means underlying the mechanistic saturation analysis of
Section~\ref{subsec:recurrent-step-scaling}.

\input{tables/ut_transition_means}

%% file: tables/lqm_low_ut_error_examples.tex
\definecolor{lqmerror}{HTML}{C0392B}
\begin{table}[t]
\centering
\scriptsize
\setlength{\tabcolsep}{3pt}
\renewcommand{\arraystretch}{1.18}
\resizebox{\columnwidth}{!}{
\begin{tabular}{@{}>{\centering\arraybackslash}m{0.10\columnwidth}>{\raggedright\arraybackslash}m{0.19\columnwidth}>{\raggedright\arraybackslash}m{0.22\columnwidth}>{\centering\arraybackslash}p{0.035\columnwidth}>{\raggedright\arraybackslash}p{0.36\columnwidth}>{\centering\arraybackslash}p{0.075\columnwidth}@{}}
\toprule
Dataset & Source & Reference & $u$ & LatentMT prediction & Error count \\
\midrule
DRT & Cruel, to deny me one ray of those adorable eyes!--how a single glance would have revived me! & \begin{CJK*}{UTF8}{gbsn}残忍啊，不让我见那可爱双眸的一丝光芒！--哪怕一瞥，也能让我精神为之一振。\end{CJK*} & \multicolumn{3}{c@{}}{\begin{tabular}{@{}>{\centering\arraybackslash}p{0.035\columnwidth}>{\raggedright\arraybackslash}p{0.36\columnwidth}>{\centering\arraybackslash}p{0.075\columnwidth}@{}}
1 & \begin{CJK*}{UTF8}{gbsn}\textcolor{lqmerror}{残酷地}\hspace{0.4em}\textcolor{lqmerror}{否定}\hspace{0.4em}\textcolor{lqmerror}{我那些可爱眼睛中}的一束光芒！--即便是\textcolor{lqmerror}{最微弱的一缕光芒}，\textcolor{lqmerror}{也能夺走我希望的光明}。\end{CJK*} & 5 \\
2 & \begin{CJK*}{UTF8}{gbsn}残忍，拒绝\textcolor{lqmerror}{我那令人向往的眼神}中的一缕光芒！--若有一瞥便能夺回我的生命火花！\end{CJK*} & 1 \\
3 & \begin{CJK*}{UTF8}{gbsn}残忍，竟不肯让我瞥见那可爱的眼眸中的一束光芒！--哪怕一瞥，也能让我重获生机！\end{CJK*} & 0 \\
\end{tabular}} \\
\midrule
ArzEn-MultiGenre & The thorns, \textcolor{lqmerror}{what use are they}? & \begin{arabtext}الشوك، لازمته إيه؟\end{arabtext} & \multicolumn{3}{c@{}}{\begin{tabular}{@{}>{\centering\arraybackslash}p{0.035\columnwidth}>{\raggedleft\arraybackslash}p{0.36\columnwidth}>{\centering\arraybackslash}p{0.075\columnwidth}@{}}
1 & \<إيه الشوك؟> & 1 \\
2 & \<الشوك، لازمته إيه؟> & 0 \\
3 & \<الشوك، لازمته إيه؟> & 0 \\
\end{tabular}} \\
\bottomrule
\end{tabular}
}
\caption{Examples from the linguistic error analysis where increasing recurrent computation removes annotated translation errors.
For each example, the source and reference are shown once, while the LatentMT prediction is shown for $u=1$, $u=2$, and $u=3$.
Annotated errors are shown in \textcolor{lqmerror}{color}: for the DRT example, wrong terms are chosen that do not match the semantic meaning of the source; for the ArzEn-MultiGenre example, the phrase ``what use are they'' is omitted from the low-$u$ prediction.
Both examples have no annotated errors for $u\geq3$.}
\label{tab:lqm-low-ut-error-examples}
\end{table}

%% file: tables/ut_transition_means.tex
\begin{table}[!ht]
\centering
\footnotesize
\begin{tabular}{@{}lrrrrr@{}}
\toprule
$u$ & $\boldsymbol{1\!\rightarrow\!2}$ & $\boldsymbol{2\!\rightarrow\!3}$ & $\boldsymbol{3\!\rightarrow\!4}$ & $\boldsymbol{4\!\rightarrow\!5}$ & $\boldsymbol{5\!\rightarrow\!6}$ \\
\midrule
\multicolumn{6}{@{}l@{}}{\textit{DRT --- Cosine distance}} \\
2 & 0.2317 &  &  &  &  \\
3 & 0.2210 & 0.0954 &  &  &  \\
4 & 0.1923 & 0.1048 & 0.0431 &  &  \\
5 & 0.1744 & 0.0917 & 0.0620 & 0.0218 &  \\
6 & 0.1585 & 0.0784 & 0.0538 & 0.0350 & 0.0141 \\
\addlinespace
\multicolumn{6}{@{}l@{}}{\textit{DRT --- Top-1 flip rate}} \\
2 & 0.2063 &  &  &  &  \\
3 & 0.2768 & 0.1113 &  &  &  \\
4 & 0.3123 & 0.1480 & 0.0710 &  &  \\
5 & 0.3224 & 0.1633 & 0.1032 & 0.0489 &  \\
6 & 0.3201 & 0.1706 & 0.1131 & 0.0700 & 0.0388 \\
\addlinespace
\multicolumn{6}{@{}l@{}}{\textit{ArzEn-MultiGenre --- Cosine distance}} \\
2 & 0.4657 &  &  &  &  \\
3 & 0.3157 & 0.2505 &  &  &  \\
4 & 0.2767 & 0.1446 & 0.1576 &  &  \\
5 & 0.2509 & 0.1151 & 0.0950 & 0.1041 &  \\
6 & 0.2378 & 0.1192 & 0.0685 & 0.0649 & 0.0922 \\
\addlinespace
\multicolumn{6}{@{}l@{}}{\textit{ArzEn-MultiGenre --- Top-1 flip rate}} \\
2 & 0.4459 &  &  &  &  \\
3 & 0.4678 & 0.2107 &  &  &  \\
4 & 0.4654 & 0.2511 & 0.1219 &  &  \\
5 & 0.4656 & 0.2780 & 0.1809 & 0.0800 &  \\
6 & 0.4193 & 0.2631 & 0.1738 & 0.1056 & 0.0623 \\
\bottomrule
\end{tabular}
\caption{Mean mechanistic change at each recurrent transition, per recurrent-depth checkpoint $u$ (rows): cosine distance of the last hidden state and top-1 flip rate after the language-model head, averaged over generated tokens and examples. Each row $u$ shows that checkpoint's transitions $1\!\rightarrow\!2,\dots,(u{-}1)\!\rightarrow\!u$; blank cells do not exist for that checkpoint. Smaller values indicate less step-to-step change.}
\label{tab:ut-transition-means}
\end{table}